\pdfoutput=1
\documentclass[letterpaper]{article} 
\usepackage{aaai24}  
\usepackage{times}  
\usepackage{helvet}  
\usepackage{courier}  
\usepackage[hyphens]{url}  
\usepackage{graphicx} 
\usepackage{natbib}  
\usepackage{caption} 
\usepackage{algorithm}
\usepackage{algorithmic}
\usepackage{multicol}
\usepackage{multirow}
\usepackage{amsfonts}
\usepackage{amsmath}
\usepackage{amssymb}
\usepackage{pifont}
\usepackage{dsfont}
\newcommand{\ve}[1]{\mathbf{#1}}
\newcommand{\vv}[1]{\mbox{\boldmath $#1$}}

\usepackage{newfloat}
\usepackage{listings}
\DeclareCaptionStyle{ruled}{labelfont=normalfont,labelsep=colon,strut=off} 
\floatstyle{ruled}
\newfloat{listing}{tb}{lst}{}
\floatname{listing}{Listing}
\title{TagFog: Textual Anchor Guidance and Fake Outlier Generation for Visual Out-of-Distribution Detection}
\author{
    Jiankang Chen\textsuperscript{\rm 1,3}, 
    Tong Zhang\textsuperscript{\rm 2}, 
    Wei-Shi Zheng\textsuperscript{\rm 1,3}, 
    Ruixuan Wang\textsuperscript{\rm 1,2,3}\thanks{Corresponding author}
}

\affiliations{
    \textsuperscript{\rm 1}School of Computer Science and Engineering, Sun Yat-sen University, Guangzhou, China\\
    \textsuperscript{\rm 2}Peng Cheng Laboratory, Shenzhen, China\\
    \textsuperscript{\rm 3}Key Laboratory of Machine Intelligence and Advanced Computing, MOE, Guangzhou, China\\
    chenjk36@mail2.sysu.edu.cn, zhangt02@pcl.ac.cn, wszheng@ieee.org, wangruix5@mail.sysu.edu.cn
}

\usepackage{bibentry}
\begin{document}

\maketitle

\begin{abstract}
Out-of-distribution (OOD) detection is crucial in many real-world applications. However, intelligent models are often trained solely on in-distribution (ID) data, leading to overconfidence when misclassifying OOD data as ID classes.  In this study, we propose a new learning framework which leverage simple Jigsaw-based fake OOD data and rich semantic embeddings (`anchors') from the ChatGPT description of ID knowledge to help guide the training of the image encoder. The learning framework can be flexibly combined with existing post-hoc approaches to OOD detection, and extensive empirical evaluations on multiple OOD detection benchmarks demonstrate that rich textual representation of ID knowledge and fake OOD knowledge can well help train a visual encoder for OOD detection. With the learning framework, new state-of-the-art performance was achieved on all the benchmarks. The code is available at \url{https://github.com/Cverchen/TagFog}.
\end{abstract}

\section{Introduction}

When deploying well-trained AI models in real-world applications, AI models often encounter samples which are different from the distributions of training data~\cite{baseline,baseline2,baseline3}. Such out-of-distribution (OOD) samples are often from unknown classes which did not appear during model training. Mis-classifying OOD samples into previously learned in-distribution (ID) classes could lead to serious consequences such as in the autonomous driving and the intelligent healthcare applications. Therefore, it is a desired ability for the AI model to accurately detect whether a new data is an OOD sample or from one of previously learned classes.

Various approaches have been developed for solving the OOD detection problem. Most approaches train a decent classifier on ID data, then use the feature output of the penultimate layer, logits output of the classifier, or the softmax probability vector output to design a score function \cite{MSP,Maha,Energy,Vim}. The defined score is typically lower for OOD data compared to ID data. However, training only on ID data can cause overconfidence, with models assigning high confidence to unseen OOD data~\cite{LogitNorm}.

\begin{figure}[t]
\centering
\includegraphics[height=0.6\columnwidth, width=0.9\columnwidth, keepaspectratio]{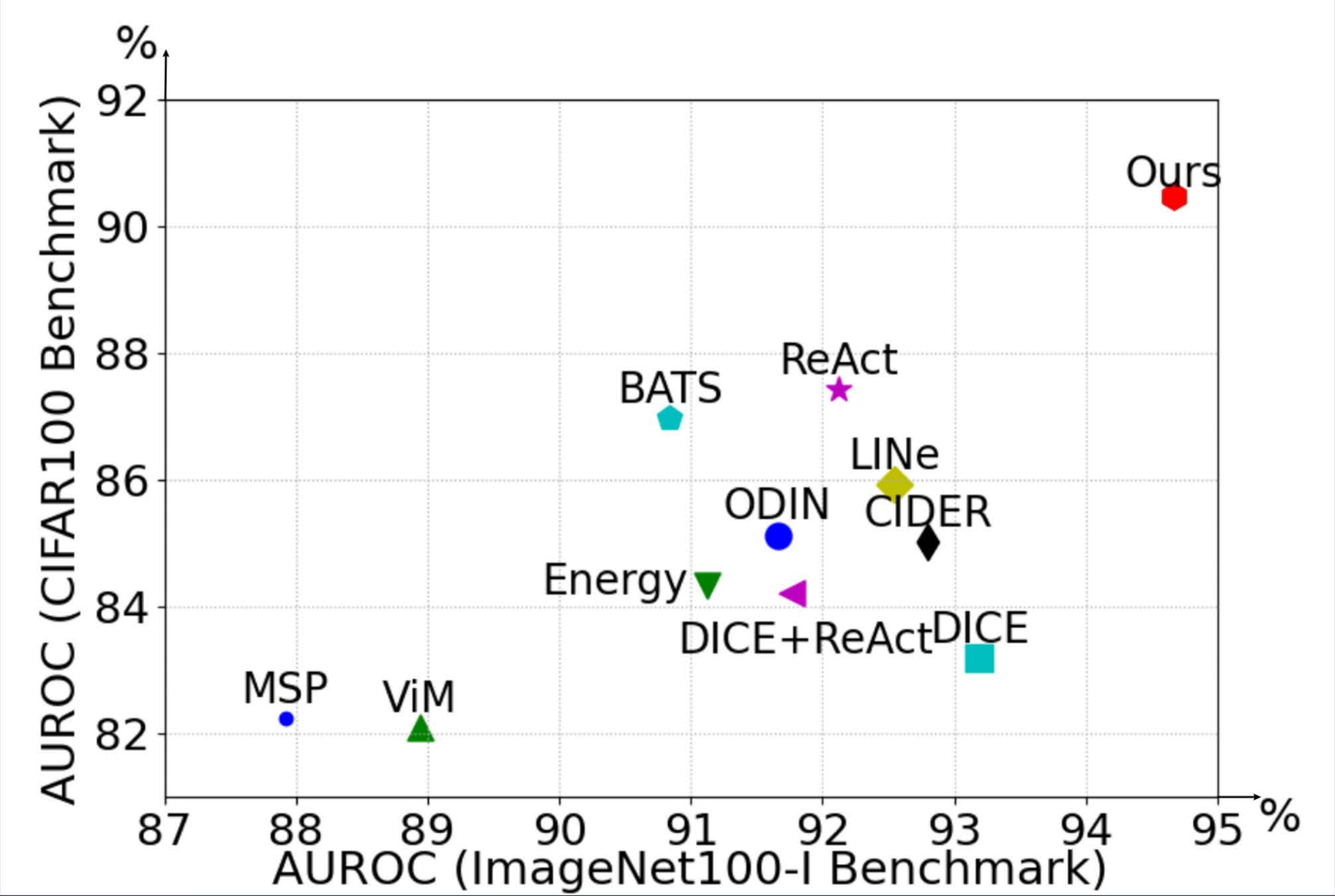}
\caption{OOD detection performance of different methods
on the CIFAR100 and ImageNet100-I benchmarks.}
\label{fig1}
\end{figure}

If certain OOD samples are available during training, the model will gain knowledge of data that are characterized differently from the ID data. This would help the model better identify OOD data later. 
However, obtaining OOD data in certain real-world applications is often time-consuming or costly. Based on above considerations, researchers have proposed various strategies to generate fake OOD data from available ID training data.  
One way is to use generative adversarial networks (GANs)~\cite{gan} for generating fake OOD samples based on available ID data~\cite{gan1,gan4,gan6,gan3}. However, 
it is often challenging for GANs to generate expected OOD samples due to the unstable training and difficulty in generating realistic OOD samples based on only ID samples~\cite{gan5,gan2}. 
Instead of generating fake OOD data in the input space as by GANs, OOD knowledge may also be gained from the feature space. For example, VOS~\cite{VOS} synthesizes virtual OOD features from the low likelihood regions of ID data in the high-level feature space to help improve OOD detection. This method assumes a strict Gaussian distribution for ID data, which is often unrealistic.
Different from directly generating OOD data to gain OOD knowledge, the pre-trained large vision-language model CLIP~\cite{CLIP} has recently been used to help OOD detection considering that much knowledge, including OOD knowledge, has been learned by the CLIP model~\cite{CLIP_morelabel,CLIP_lablegenerate,MCM}. However, this approach requires unrealistic OOD data labels and the pre-trained visual encoder during OOD detection. 

In this study, we propose a simple yet effective learning framework \textbf{TagFog} (\textbf{T}extual \textbf{a}nchor \textbf{g}uidance and \textbf{F}ake \textbf{o}utlier \textbf{g}eneration) to train a visual model for OOD detection based on a simple fake OOD generation strategy and a CLIP-based textual guidance with the description of ID knowledge from {the ChatGPT~\cite{gpt}}. The fake OOD data are generated offline by the simple Jigsaw transformations~\cite{jigsaw} on training ID images such that fake OOD data are similar to corresponding ID data at patch level, but differently at the image level. In this way, fake OOD data would contain knowledge which is semantically shifted from that of ID data, and therefore can be considered as challenging OOD samples to help the model better discriminate between ID and real OOD data. On the other hand, the textual description of each ID class from the ChatGPT contains richer information compared to the solely ID class name, and therefore CLIP's embedding of the ChatGPT description would contain semantically more information about ID knowledge. In this study,  the CLIP's textual embeddings of ID knowledge as anchors are used to guide the training of the image encoder with contrastive learning, such that the image encoder can learn to extract richer and more compact representations from images.
Our approach demonstrates the power of using textual guidance and fake OOD data for OOD detection, as supported by extensive empirical evaluations on multiple OOD detection benchmarks.
The main contributions are summarized below.
\begin{itemize}
    \item A simple yet effective learning framework which uses fake OOD data and rich textual embeddings of ID classes to help train a better image encoder. Notably, the framework can be flexibly fused with many existing methods.
    \item The first usage of ChatGPT for more informative and semantic embeddings of ID knowledge which are used to guide training of the image encoder for OOD detection.
    \item Extensive experimental evaluations on multiple OOD detection benchmarks, with state-of-the-art performance obtained from our approach. 
\end{itemize}

\section{Preliminaries}
\textbf{Out-of-Distribution Detection.} Suppose $K$ classes of training data are available to train a classifier. In the OOD detection task, the classifier is expected to decide whether a new data belongs to one of the $K$ classes or from any unseen class. Data from the $K$ classes are in-distribution (ID), while data from any unseen class are out-of-distribution. 
OOD detection can be viewed as a binary classification task. 
Usually, a scoring function $\mathcal{S}_{\lambda}$ based on the classifier's output at certain layer is designed for OOD detection, where $\lambda$ is the threshold.  
For any new data as input to the classifier, when the score is above the threshold $\lambda$, the input data is determined as ID. Otherwise, the input is considered as OOD.

\noindent\textbf{Pre-trained Vision-Language Model CLIP.}
Knowledge learned only from images  is limited. In contrast, visual-language contrastive representation learning achieves much better performance on downstream tasks. A representative vision-language model is CLIP which includes a text encoder and an image encoder~\cite{CLIP}. 400 million image-text pairs on websites are crawled for training of the CLIP model, based on the contrastive InfoNCE loss by maximizing the similarity between matched image-text pairs and minimizing the similarity for mismatched pairs. Both the text encoder and the image encoder of the well-trained CLIP are expected to encode semantically rich information from the corresponding text and image input. 

\begin{figure*}[th]
    \centering
    \includegraphics[width=0.95\textwidth, height=0.45\textwidth]{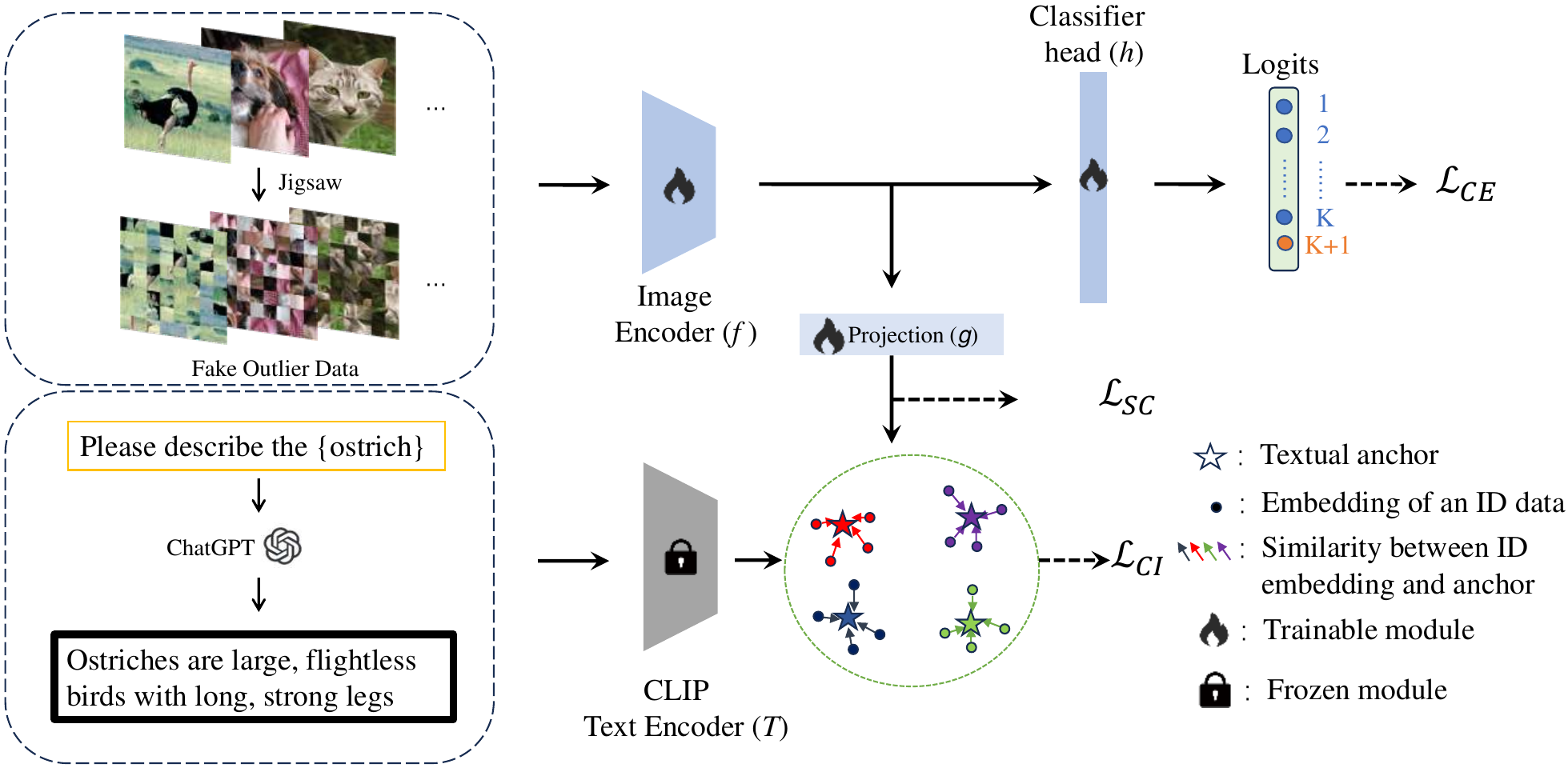} 
    \caption{Overview of the proposed learning framework TagFog for OOD detection. Upper part: fake OOD data are generated based on the Jigsaw strategy and, together with the ID data, used to train the image encoder $f$ and the classifier head $h$. 
   Lower part: the description of each ID class from ChatGPT is fed to the pretrained and fixed CLIP's Text Encoder to obtain the semantic embedding as anchor for the ID class. The anchors are used to guide the training of the image encoder based on the contrastive loss $\mathcal{L}_{CI}$ and $\mathcal{L}_{SC}$. 
    }
    \label{fig:framework}
\end{figure*}

\section{Method}

\textbf{Overview.} \ \  Our framework TagFog is illustrated in Figure~\ref{fig:framework}. The framework mainly contains two parts. The first part (i.e., upper part of Figure~\ref{fig:framework}) makes use of fake OOD data to help train a $(K+1)$-class classifier, where the fake OOD data are generated based on the training data of $K$ ID classes and expected to help the classifier better discriminate between ID data and real OOD data during inference.  The second part (i.e., lower part of Figure~\ref{fig:framework}) novelly applies ChatGPT to generate descriptive text for each ID class, and such text is then fed into the pretrained and fixed CLIP's Text Encoder to create semantic embedding for the corresponding ID class. 
The embedding serves as the anchor for all training data of the corresponding ID class, and the anchors of all $K$ ID classes are used to help guide the training of the Image Encoder based on the alignment between the associated anchor and the projection of the Image Encoder output for each ID data. {In addition, the idea of SupCon~\cite{SupCon} is utilized to perform supervised contrast learning on all projected ID and fake OOD embedding vectors.} With the help of the ChatGPT-generated semantic guiding and the fake OOD-involved contrastive learning and classifier training, the Image Encoder is expected to learn to generate compact ID feature representations while leaving much spare regions for OOD data in the feature space.

\noindent\textbf{Fake Outlier Generation (FOG).} Usage of OOD data during classifier training has been shown helpful to improve OOD detection performance~\cite{gan2,gan3,oe_1,oe_2}. However, most OOD detection methods~\cite{MSP,ODIN,CIDER} train a classifier based only on ID data and therefore the classifier would not contain any knowledge of OOD data.
On the other hand, 
approaches using fake OOD data during classifier training  are either based on unstable GAN models~\cite{gan1} or make overly constrained  feature space assumptions~\cite{VOS}, which often includes a complicated fake OOD generation process and may not work effectively in various real applications. 

In this study, we propose a simple yet effective fake OOD data generation strategy based on the Jigsaw technique. Specifically, for each ID training image, the image is  divided into multiple image patches which are then randomly shuffled and rearranged to form a new image. The synthesized new image is considered as a fake OOD data for model training. Because the Jigsaw process disrupts the overall structure and contextual information in the original image, the original semantic information is altered, resulting in semantic offset from the original image. In other words, the semantics of object(s) of interest in the original image is largely distorted due to the shuffling of image patches. Since some image patches in the fake OOD data contain parts of ID objects, the fake OOD data would contain information which is partially similar to  but semantically shifted from the ID image, thus can be served as challenging OOD data during model training. 
In addition, the image patches containing only background information appear both in the fake OOD data and the corresponding ID data. In order to differentiate the fake OOD data from the ID data, the classifier would need to focus on learning from the (foreground) object regions, thus alleviating the overconfidence issue of mis-classifying OOD data as an ID class due to unique background in the ID data~\cite{background}. Note that the Jigsaw technique has been used recently in the FeatureNorm method~\cite{FeatureNorm}, not for model training, but for layer selection after the model is trained as usual. In contrast, here the Jigsaw-based fake OOD data are used for model training.

\noindent \textbf{Textual Anchor Guidance (TAG).} More compact and semantically informative representations are beneficial for OOD detection~\cite{MCM, CIDER}. 
In order to help the classifier learn to extract more relevant semantic information from input images, here we utilize the large-scale language model ChatGPT and the large-scale vision-language model CLIP to help guide the training of the image classifier. Specifically, ChatGPT is used to generate semantically rich description for each ID class (Table~\ref{tab:chatgpt}),\begin{table}[tbh]
\centering  
\begin{tabular}{l|l}
\hline
ID class &\ \ \ \ \ \multirow{2}{*}{Textual description from ChatGPT} \\ 
\ \ \ name & \\
\hline
\ \ bee & \ \ \ insect, black and yellow stripes, ...\\  
\ \ cloud & \ \ \ visible mass of condensed water vapor,... \\   
\ \ cup & \ \ \ small container for drinking, made of... \\
\ \ sea & \ \ \ large body of saltwater covering most of...\\
\hline
\end{tabular}
\caption{Demonstrative textual descriptions of ID classes. The textual description of each ID class is obtained by asking ChatGPT ``Please describe the  \{ID class name\}''.} 
\label{tab:chatgpt}
\end{table} and the textual description is then fed to the CLIP's Text Encoder to obtain the semantic embedding (namely `anchor') of the ID class. The anchors are expected to contain more semantic information than the textual embedding of solely ID class names. To align with the associated anchor, the visual encoder's outputs for each input image is projected into the semantic embedding space, and both the image encoder and the projection module will be trained such that the projected visual embedding for each input image is as close to the associated textual anchor as possible. 

\noindent \textbf{Model Training.} The image encoder $f$, the classifier head $h$, and the projection module $g$ need to be trained. Note that the output of the classifier head is $(K+1)$-dimensional, with $K$ outputs for ID classes and one  output for OOD class. As usual, the cross-entropy loss function $\mathcal{L}_{CE}$ is used to train the encoder and the classifier head, 
\begin{equation}
\mathcal{L}_{CE} = - \frac{1}{N+M}\sum_{i=1}^{N+M} \sum_{k=1}^{K+1} y_{i,k} \log(\hat{y}_{i,k})
\end{equation}
where $N$ and $M$ are respectively the number of all ID training images and fake OOD images, $\hat{y}_{i,k}$ is the output probability for the $i$-th training image belonging to the $k$-th class, and $y_{i,k}$ is the corresponding ground-truth output ($0$ or $1$). 

For the CLIP-based textual guidance, denote by $\ve{t}_k$ the textual description from the ChatGPT for the $k$-th ID class, and $\vv{\mu}_k = \mathcal{T}(\ve{t}_k)$ the corresponding anchor vector from the output of the CLIP's text encoder for the $k$-th ID class. In order to attract the projected visual embedding for each input image close to the associate anchor, the contrastive loss $\mathcal{L}_{CI}$ is designed as below
\begin{small}
\begin{equation}
    \mathcal{L}_{CI}=-\frac{1}{N}\sum_{n=1}^{N}\sum_{k=1}^{K} {\mathds{1}({{y}_{n,k} \neq 0}) \cdot \log \frac{\exp (s(\ve{z}_{n}, \vv{\mu}_{k}) / \tau)}{\sum_{j=1}^{ K} \exp (s(\ve{z}_{n}, \vv{\mu}_{j}) / \tau)}} \,,
\end{equation}
\end{small}where $\ve{z}_{n} = g(f(\ve{x}_n))$ is the projected visual embedding for the input ID image $\ve{x}_n$, and $s(\ve{z}_{n}, \vv{\mu}_{k})$ represents the cosine similarity between the two embeddings $\ve{z}_{n}$ and $\vv{\mu}_{k}$. $\mathds{1}(\cdot)$ is the indicator function and $\tau$ is the temperature scaling factor. By minimizing $\mathcal{L}_{CI}$, the image encoder $f$ and the projection module $g$ (here with structure Linear-BN-ReLU-Linear) will be trained such that the projected visual embeddings of the same ID class will be close to the associated anchor, therefore helping the image encoder extract more compact and semantically informative features.

To further differentiate fake OOD images from anchor-guided ID images, the supervised contrastive loss $\mathcal{L}_{SC}$ is utilized following the idea of SupCon~\cite{SupCon},
\begin{small}
\begin{equation}
    \mathcal{L}_{{SC}}=-\frac{1}{S}\sum_{i=1}^{S} \frac{1}{|P(i)|}\sum_{p \in P(i)} \log \frac{\exp \left(s(\ve{z}_{i}, \ve{z}_{p} )/ \tau' \right)}{\sum_{a \in A(i)} \exp \left(s(\ve{z}_{i}, \ve{z}_{a}) / \tau' \right)} \,,
\end{equation}
\end{small}{where 
$S = N+M$,} $\mathcal{\textit{A(i)}}$ represents all the sample indices in the mini-batch that includes the sample with index $\mathcal{\textit{i}}$, and $\mathcal{\textit{P(i)}}$ is the subset 
of $\mathcal{\textit{A(i)}}$ in which all the corresponding samples share the same class label as the that of the sample with index $i$. 
$\tau'$ is the temperature scaling factor. 

Overall, the image encoder $f$, the classifier head $h$, and the projection module $g$ can be trained by minimizing the combined loss function $\mathcal{L}$, 
\begin{equation}
    \mathcal{L} = \mathcal{L}_{CE} + \lambda_1 \mathcal{L}_{CI} + \lambda_2 \mathcal{L}_{SC} \,,
\end{equation}
with coefficients $\lambda_1$ and $\lambda_2$ balancing the three loss terms. 

\noindent\textbf{Model Inference.} Once the model is well trained, the image encoder together with the classifier head is used to detect whether a new image is OOD or not. Since our method focuses on model training, any post-hoc OOD detection strategy can be utilized during model inference. By default here the recently proposed post-hoc method ReAct~\cite{React} is used for OOD detection. Note that only the logit values of the $K$ ID classes are used to calculate the ReAct score, although the output of the fake OOD class in the classifier head may also be investigated to further improve the OOD detection performance.

\begin{table*}[ht]
\centering
\setlength{\tabcolsep}{4pt}
\small
\begin{tabular}{ccccccccccccccccc}
\hline
\multirow{3}{*}{\begin{tabular}[c]{@{}c@{}}{ID Dataset}\end{tabular}} 
&\multirow{3}{*}{Method} & \multicolumn{12}{c}{OOD Datasets} & \multicolumn{2}{c}{\multirow{2}{*}{Average}} \\ \cline{3-14} 
&  & \multicolumn{2}{c}{SVHN} & \multicolumn{2}{c}{LSUN-R} & \multicolumn{2}{c}{LSUN-C} & \multicolumn{2}{c}{iSUN} & \multicolumn{2}{c}{Textures} & \multicolumn{2}{c}{Places365} \\
& & F$\downarrow$ & A$\uparrow$ & F$\downarrow$ & A$\uparrow$ & F$\downarrow$ & A$\uparrow$ & F$\downarrow$ & A$\uparrow$ & F$\downarrow$ & A$\uparrow$ & F$\downarrow$ & A$\uparrow$ & F$\downarrow$ & A$\uparrow$ \\ \hline
\multirow{14}{*}{\begin{tabular}[c]{@{}c@{}}CIFAR10\end{tabular}} 
& MSP & 61.22 & 86.99 & 41.62 & 93.84 & 34.30 & 95.40 & 43.14 & 93.21 & 53.40 & 90.19 & 54.51 & 88.74 & 48.03 & 91.40 \\
& Mahalanobis & 67.25 & 89.51 & 48.37 & 92.38 & 91.65 & 74.55 & 44.24 & 92.68 & 45.92 & 91.96 & 66.11 & 85.79 & 60.59 & 87.81 \\
& ODIN & 53.56 & 77.48 & 17.31 & 94.63 & 13.64 & 96.09 & 19.87 & 93.55 & 46.65 & 80.85 & 49.72 & 79.92 & 33.46 & 87.09 \\
& Energy & 41.25 & 87.69 & 24.19 & 95.01 & 11.37 & 97.63 & 26.40 & 94.16 & 42.52 & 89.10 & 40.04 & 88.71 & 30.96 & 92.05 \\
& ViM & 53.75 & 88.67 & 34.17 & 94.34 & 82.31 & 87.18 & 31.41 & 94.25 & 36.15 & 92.83 & 49.64 & 88.86 & 47.90 & 91.02 \\
& DICE & 36.42 & 91.46 & 31.57 & 93.77 & 7.10 & 98.67 & 36.94 & 92.05 & 47.02 & 88.41 & 46.74 & 86.05 & 34.30 & 91.73 \\ 
& BATS & 41.42 & 87.84 & 24.17 & 95.02 & 11.35 & 97.63 & 26.36 & 94.16 & 42.13 & 89.29 & 40.04 & 88.71 & 30.91 & 92.11 \\
& ReAct & 43.19 & 87.56 & 24.82 & 95.12 & 12.23 & 97.53 & 26.90 & 94.31 & 41.95 & 90.02 & 40.78 & 89.00 & 31.65 & 92.26 \\ 
& DICE+ReAct & 36.90 & 91.31 & 31.59 & 93.71 & 7.29 & 98.64 & 37.15 & 92.10 & 46.76 & 88.61 & 46.76 & 86.12 & 34.41 & 91.75 \\
& FeatureNorm & \textbf{2.37} & 99.45 & 33.42 & 94.71 & \textbf{0.10} & \textbf{99.93} & 27.01 & 95.65 & 23.03 & 95.65 & 58.96 & 87.95 & 24.14 & 95.55\\
& LINe & 45.38 & 87.96 & 39.25 & 92.61 & 9.75 & 98.19 & 41.52 & 91.74 & 58.37 & 84.14 & 53.02 & 85.70 & 41.22 & 90.06 \\
& VOS & 35.73 & 93.74 & 25.54 & 95.29 & 18.47 & 96.55 & 30.17 & 94.16 &44.16 & 90.07 & 44.18 & 88.13 & 33.04 & 92.99\\
& LogitNorm & 12.68 & 97.75 & 15.29 & 97.45 & 0.53 & 99.82 & 15.36 & 97.43 & 31.56 & 94.09 & 32.31 & 93.92 & 17.96 & 96.75 \\
& CIDER & 2.89 & \textbf{99.72} & 23.13 & 96.28 & 5.45 & 99.01 & 20.21 & 96.64 & \textbf{12.33} & 96.85 & \textbf{23.88} & 94.09 & 14.64 & 97.10 \\
& \textbf{TagFog (Ours)}& 6.19 & 98.75 & \textbf{6.50} & \textbf{98.74} & 2.12 & 99.43 & \textbf{6.36} & \textbf{98.75} & 16.13 & \textbf{97.12} & 25.14 & \textbf{95.14} & \textbf{10.41} & \textbf{97.99} \\ 
 \hline
 \multirow{14}{*}{\begin{tabular}[c]{@{}c@{}}CIFAR100\end{tabular}} 
 & MSP & 69.74 & 84.73 & 66.89 & 85.65 & 77.08 & 81.83 & 69.40 & 84.77 & 80.08 & 77.65 & 78.38 & 78.81 & 73.60 & 82.24 \\
 & Mahalanobis & 92.62 & 66.80 & 89.00 & 68.46 & 98.83 & 49.58 & 88.45 & 68.44 & 72.68 & 74.57 & 92.87 & 63.26 & 89.07 & 65.18 \\
 & ODIN & 79.74 & 81.40 & 37.63 & 93.21 & 72.66 & 85.93 & 39.59 & 92.58 & 73.07 & 80.42 & 80.39 & 77.22 & 63.85 & 85.13 \\
 & Energy & 68.90 & 87.66 & 59.71 & 88.58 & 73.21 & 84.46 & 64.03 & 87.50 & 79.61 & 78.22 & 77.74 & 79.64 & 70.53 & 84.34 \\
 & ViM & 73.70 & 84.45 & 61.30 & 88.05 & 92.76 & 69.87 & 61.92 & 87.34 & 57.93 & 86.31 & 81.01 & 76.54 & 71.43 & 82.09 \\
 & DICE & 53.60 & 90.22 & 79.84 & 81.17 & 40.03 & 92.52 & 79.79 & 80.96 & 78.65 & 77.46 & 82.31 & 76.76 & 69.04 & 83.18 \\ 
 & BATS & 62.05 & 89.31 & 50.38 & 91.21 & 73.70 & 84.55 & 55.97 & 90.30 & 72.93 & 84.50 & 72.61 & 82.03 & 64.61 & 86.98 \\ 
 & ReAct & 58.24 & 90.02 & 50.82 & 90.98 & 70.70 & 85.75 & 55.91 & 90.18 & 70.85 & 85.39 & \textbf{71.85} & \textbf{82.25} & 63.06 & 87.43 \\ 
 & DICE+ReAct & 48.20 & 91.19 & 84.18 & 78.79 & 32.05 & 93.71 & 82.23 & 79.65 & 66.74 & 83.96 & 80.28 & 77.96 & 65.61 & 84.21\\
 & FeatureNorm & \textbf{15.98} & \textbf{96.59} & 96.57 & 61.80 & \textbf{4.56} & \textbf{98.95} & 93.56 & 65.15 & 51.67 & 83.54 & 93.61 & 56.83 & 59.33 & 77.07 \\
 & LINe & 52.02 & 91.01 & 65.66 & 86.87 & 47.76 & 91.23 & 69.27 & 85.90 & 71.22 & 83.37 & 80.90 & 77.21 & 64.47 & 85.93\\
 & VOS & 78.36 & 80.58 & 69.77 & 84.77 & 77.38 & 83.61 & 69.65 & 85.48 & 76.60 & 80.58 & 80.47 & 77.57 & 75.37 & 81.96\\
 & LogitNorm & 51.34 & 91.79 & 88.80 & 78.67 & 6.82 & 98.70 & 90.16 & 75.55 & 77.02 & 77.52 & 77.79 &79.56 & 65.32 & 83.63\\
 & CIDER & 31.36 & 93.47 & 80.39 & 81.54 & 43.68 & 89.45 & 78.23 & 81.33 & \textbf{35.51} & \textbf{91.70} & 82.80 & 72.71 & 58.66 & 85.03\\
 & \textbf{TagFog (Ours)} & 37.88 & 92.77 & \textbf{35.45} & \textbf{93.46} & 13.94 & 97.46 & \textbf{35.99} & \textbf{93.10} & 66.74 & 86.88 & 76.00 & 79.13 & \textbf{44.33} & \textbf{90.47} \\ 
 \hline
\end{tabular}%
\caption{OOD detection performance on the CIFAR10 and the CIFAR100(ID) benchmarks with model backbone ResNet18. $\uparrow$ indicates that larger values are better and $\downarrow$ indicates that smaller values are better. All values are percentages.}
\label{tab:cifar10}\end{table*}
\begin{table*}[bht]
\centering
\setlength{\tabcolsep}{6pt}
\normalsize
\begin{tabular}{cccccccccccccccc}
\hline
\multirow{2}{*}{\begin{tabular}[c]{@{}c@{}}ID Dataset\end{tabular}} & \multirow{2}{*}{Metrics} & \multicolumn{8}{c}{Method} \\ \cline{3-10}
& & MSP & Mahalanobis & ODIN & DICE & VIM & Energy & BATS & ReAct\\ \hline
\multicolumn{1}{c|}{\multirow{6}{*}{CIFAR10}} & \multicolumn{1}{c|}{F$\downarrow$}& 40.95 & 55.69 & 33.06 & 36.63 & 38.35 & 26.69 & 29.85 & 27.76\\ 
\multicolumn{1}{c|}{}& \multicolumn{1}{c|}{A$\uparrow$}& 92.09 & 91.99 & 89.64 & 90.72 &93.10  & 93.75 & 93.17 & 93.60\\ \cline{3-10}
\multicolumn{1}{c|}{}&\multicolumn{1}{c|}{} & DICE+ReAct& FeatureNorm & LINe & VOS & LogitNorm &CIDER &  \multicolumn{2}{c}{\textbf{TagFog (Ours)}} \\ \cline{3-10} 
 \multicolumn{1}{c|}{}& \multicolumn{1}{c|}{F$\downarrow$}& 36.85& 29.76 & 42.97&27.01 & 16.95& 16.10  & \multicolumn{2}{c}{\textbf{11.17}}\\ 
\multicolumn{1}{c|}{}& \multicolumn{1}{c|}{A$\uparrow$}& 93.29 & 90.72 &  89.42& 94.04& 96.93& 97.25  &\multicolumn{2}{c}{\textbf{97.72}}\\ \cline{3-10} \hline
 
\multicolumn{1}{c|}{\multirow{6}{*}{CIFAR100}} & \multicolumn{1}{c|}{} & MSP & Mahalanobis & ODIN & DICE & VIM & Energy & BATS & ReAct\\ \cline{3-10}
\multicolumn{1}{c|}{}& \multicolumn{1}{c|}{F$\downarrow$} &78.29 & 93.86 & 64.01 & 68.14 & 61.51 &59.60  & 69.41 & 56.73 \\
\multicolumn{1}{c|}{}& \multicolumn{1}{c|}{A$\uparrow$}& 79.25 & 55.21 & 83.44 & 83.53 & 85.00 & 83.64 & 87.52 & 88.30  \\ \cline{3-10}
\multicolumn{1}{c|}{}&\multicolumn{1}{c|}{} & DICE+ReAct& FeatureNorm & LINe & VOS & LogitNorm &CIDER &  \multicolumn{2}{c}{\textbf{TagFog (Ours)}} \\ \cline{3-10} 
\multicolumn{1}{c|}{}& \multicolumn{1}{c|}{F$\downarrow$}& 50.56& 65.61&  54.52 &80.53 & 70.81 & 50.66 & \multicolumn{2}{c}{\textbf{45.28}}\\ 
\multicolumn{1}{c|}{}& \multicolumn{1}{c|}{A$\uparrow$}& 84.22 &80.12 &  86.32 & 79.03 & 79.44&86.70 & \multicolumn{2}{c}{\textbf{90.20}}\\ \cline{3-10} \hline
\end{tabular}
\caption{OOD detection performance on the CIFAR10 and the CIFAR100 benchmarks with model backbone ResNet34. Values are average percentages over six OOD datasets.}
\label{tab:cifar}
\end{table*}\begin{table*}[!bht]
\centering
\setlength{\tabcolsep}{6pt}
\normalsize
\begin{tabular}{ccccccccccccccccc}
\hline
\multirow{2}{*}{\begin{tabular}[c]{@{}c@{}}Model\end{tabular}} & \multirow{2}{*}{Metrics} & \multicolumn{8}{c}{Method} \\ \cline{3-10}
& & MSP &ODIN &  Mahalanobis & Energy & GradNorm & ViM & KNN & BATS\\ \hline
\multicolumn{1}{c|}{\multirow{6}{*}{ResNet50}} & \multicolumn{1}{c|}{F$\downarrow$}& 58.54 & 42.43 & 80.60 & 46.72 & 41.94 & 57.97 & 40.04 & 44.81\\ 
\multicolumn{1}{c|}{}& \multicolumn{1}{c|}{A$\uparrow$}& 87.92 & 91.66 & 60.74 & 91.12 & 89.09 & 88.94 & 90.68 & 90.84\\ \cline{3-10}
\multicolumn{1}{c|}{}&\multicolumn{1}{c|}{} & DICE & ReAct & DICE+ReAct & FeatureNorm & LINe & CIDER &  \multicolumn{2}{c}{\textbf{TagFog (Ours)}} \\ \cline{3-10} 
 \multicolumn{1}{c|}{}& \multicolumn{1}{c|}{F$\downarrow$}& 32.63& 39.85 & 40.51 & 61.33& 34.66 & 39.74 &\multicolumn{2}{c}{\textbf{29.50}}\\ 
\multicolumn{1}{c|}{}& \multicolumn{1}{c|}{A$\uparrow$}& 93.19 & 92.12& 91.77 & 84.12& 92.55 & 92.80  &\multicolumn{2}{c}{\textbf{94.67}}\\ \cline{3-10} \hline
 
\multicolumn{1}{c|}{\multirow{6}{*}{ResNet101}} & \multicolumn{1}{c|}{} & MSP &ODIN &  Mahalanobis & Energy & GradNorm & ViM & KNN & BATS\\ \cline{3-10}
\multicolumn{1}{c|}{}& \multicolumn{1}{c|}{F$\downarrow$} & 55.56 & 38.48 & 77.85 & 43.82 & 43.57 & 51.21 & 39.12 & 39.32 \\
\multicolumn{1}{c|}{}& \multicolumn{1}{c|}{A$\uparrow$}& 88.74 & 92.20 & 66.35 & 91.87 & 87.88 & 90.78 & 91.60 & 91.81  \\ \cline{3-10}
\multicolumn{1}{c|}{}&\multicolumn{1}{c|}{} & DICE & ReAct & DICE+ReAct & FeatureNorm & LINe & CIDER &  \multicolumn{2}{c}{\textbf{TagFog (Ours)}} \\ \cline{3-10} 
\multicolumn{1}{c|}{}& \multicolumn{1}{c|}{F$\downarrow$}& 31.53& 39.98 & 35.52 & 48.23& 33.77 & 39.03 &\multicolumn{2}{c}{\textbf{28.53}}\\ 
\multicolumn{1}{c|}{}& \multicolumn{1}{c|}{A$\uparrow$}& 93.50 & 92.26& 92.42 & 89.23& 92.81 & 92.40  &\multicolumn{2}{c}{\textbf{94.66}}\\ \cline{3-10} \hline
\end{tabular}%
\caption{OOD detection performance on the ImageNet100-I benchmark with model backbones ResNet50 and ResNet101. Values are average percentages over four OOD datasets.}
\label{tab:imagenet100}
\end{table*}

\section{Experiments}
\subsection{Experimental Settings}
\textbf{Datasets.} Our method is evaluated on two sets of OOD detection benchmarks. {Each benchmark includes one training ID set, one test ID set and several OOD test sets.} For CIFAR~\cite{CIFAR} benchmarks, CIFAR10 and CIFAR100 were respectively used as the ID datasets, and six datasets were used as OOD test sets, including Textures \cite{textures}, SVHN~\cite{SVHN}, iSUN~\cite{iSUN},  Places365~\cite{Places}, LSUN-C~\cite{LSUN}, and LSUN-R~\cite{LSUN}. For large-scale ImageNet benchmarks, two different sets of 100 ImageNet~\cite{imagenet} classes, namely ImageNet100-I~\cite{MCM} and ImageNet100-II~\cite{ImageNet100_another}, were used as ID sets considering that both sets have been used in related literature, and four OOD test datasets, Places \cite{Places}, Textures, iNaturalist~\cite{iNaturalist}, and SUN~\cite{Sun} were used for evaluation. There are no overlapped classes between OOD datasets and corresponding ID datasets.  
Please see Supplementary Section A for more dataset details.

\noindent\textbf{Experimental details.} Following previous studies~\cite{CIDER,MSP,React}, ResNet18 \cite{ResNet} and ResNet34 were used as the model backbone on CIFAR benchmarks (please also see results with WideResNet28-10~\cite{LogitNorm} in the Supplementary Table 8), and ResNet50 and ResNet101 were used on the ImageNet100 benchmarks. To generate fake OOD data, each CIFAR image was divided into $4\times4$ patches and then randomly rearranged, and similarly each ImageNet100 image was divided  into $8\times8$ patches. For CIFAR10, two jigsaw images were generated per ID image. For CIFAR100 and ImageNet100, one jigsaw image was generated per ID image. For CLIP's Text Encoder, CLIP-L/14 based on ViT-L/14 was adopted  which has a 768-dimensional output. The projection module consists of two fully connected layers with architecture Linear-BN-ReLU-Linear and with hidden layer dimension $2\times$ the input feature dimension of the projection module. 

For the ID training sets and all fake OOD data during training, each image was randomly cropped and resized to $32\times32$ for the CIFAR sets or $224\times224$ for the ImageNet100 training set, while maintaining the aspect ratio within a scale range of 0.2 to 1. In addition, random horizontal flipping, color jittering and grayscale transformation were performed on each image. The model was trained up to 200 epochs using stochastic gradient descent with momentum 0.9 and weight decay 1e-4. The initial learning rate was 0.05, and the learning rate was warmed up from 0.01 to the initial learning rate 0.05 in the first 10 epochs when the batch size was larger than 256. The learning rate decayed by a factor of 10 at the 150-th and the 180-th epoch on CIFAR10, at the 150-th epoch on CIFAR100, and at the 100-th, 150-th, and 180-th epoch on ImageNet100. The batch size was 512 for CIFAR and 128 for ImageNet100. {The temperatures $\tau$ and $\tau'$ were set to 0.1, and $\lambda_1$ and $\lambda_2$ were both set to 1.0 for all experiments.} {During testing, only center cropping and resizing were applied on each test image.} 
More details on baselines and ReAct score are in the Supplementary Sections B and C.

\noindent\textbf{Metrics.} The evaluation metrics include the false positive rate (\textbf{F}: FPR95) of OOD samples when the true positive rate of ID samples is at 95\%, and the area under the receiver operating characteristic curve (\textbf{A}: AUROC).

\subsection{Efficacy Evaluation of the Method} 
Table~\ref{tab:cifar10} summarizes the performance of our method and numerous competitive OOD detection methods from the literature on CIFAR10 and CIFAR100. The compared post-hoc methods, which do not require model retraining, include MSP~\cite{MSP}, Mahalanobis~\cite{Maha}, ODIN~\cite{ODIN}, Energy, ViM~\cite{Vim}, DICE~\cite{DICE}, BATS~\cite{BATS}, ReAct~\cite{React}, FeatureNorm~\cite{FeatureNorm}, and LINe~\cite{LINe}. The compared methods requiring model training include VOS~\cite{VOS}, LogitNorm~\cite{LogitNorm}, and CIDER~\cite{CIDER}. We present performance on all OOD datasets as well as the average performance. 
As Table~\ref{tab:cifar10} shows, our method establishes state-of-the-art average performance on both CIFAR10 and CIFAR100 benchmarks.
For example, {our method substantially outperforms VOS which produces fake OOD data in feature space assuming a strict conditional Gaussian distribution (e.g., on the CIFAR10 benchmark, FPR95 10.41\%  vs. 33.04\%, AUROC 97.99\% vs. 92.99\%). It also surpasses the current SOTA method CIDER (e.g., on the CIFAR100 benchmark, FPR95 44.33\% vs. 58.66\%, AUROC   90.47\% vs. 85.03\%).} 
Our method achieves nearly an absolute 3\% improvement in AUROC and absolute 15\% improvement in FPR95 over the best method on the CIFAR100 benchmark. 
Similar results can be observed with the model backbone ResNet34 (Table~\ref{tab:cifar}),  where our method again outperforms all current methods. The detailed performance on six OOD datasets with the backbone ResNet34 and results with the backbone WideResNet28-10 on both benchmarks are in the Supplementary Section D.

The superior performance of our method is also confirmed on the two ImageNet100 benchmarks. Besides the aforementioned baselines whose performance on either ImageNet100 benchmark was reported in literature, the baselines KNN~\cite{KNN} and GradNorm~\cite{GradNorm} were also included for comparison. As shown in Table~\ref{tab:imagenet100}, our method with both model backbones achieves  state-of-the-art average performance on the ImageNet100-I benchmark. The detailed performance on each OOD dataset and the superior performance of our method on the other benchmark ImageNet100-II were included in Supplementary Table 10 and Table 11.
\begin{table}[tbh]
\centering
\setlength{\tabcolsep}{4pt}
\small
\begin{tabular}{cccccccc} 
\hline
 \multirow{4}{*}{Fake OOD Data}& \multirow{4}{*}{$\mathcal{L}_{CI}$}&\multirow{4}{*}{$\mathcal{L}_{SC}$} & \multicolumn{2}{c}{CIFAR100} &\multicolumn{2}{c}{ImageNet100-I} \\ 
& & & \multicolumn{2}{c}{ResNet18} & \multicolumn{2}{c}{ResNet50}\\\cline{4-5} \cline{6-7}
& & & \multicolumn{2}{c}{Average} & \multicolumn{2}{c}{Average}\\
&  & & F$\downarrow$ & A$\uparrow$& F$\downarrow$ & A$\uparrow$  \\ \hline
&  &  & 63.06 &87.43 & 39.85 & 92.12\\  
\ding{52}&  &  & 48.97 & 89.17 & 37.28 & 93.42\\  
& \ding{52} &  & 53.65 & 88.38 & 32.89 & 93.78\\  
 &  & \ding{52} & 54.87 & 88.27 & 35.76 & 93.40\\
\ding{52}& \ding{52} &  & 48.59 & 89.68 & 31.95 & 94.00\\  
&\ding{52} & \ding{52}& 48.62 & 89.82 & 32.35 &93.87\\ 
\ding{52}& &\ding{52} & 47.53 & 89.57 & 33.50 & 93.82\\
\ding{52}& \ding{52} & \ding{52}& \textbf{44.33} & \textbf{90.47} & \textbf{29.50}&\textbf{94.67}\\
\hline
\end{tabular}%
\caption{Ablation study of the proposed learning framework.} 
\label{tab:ablation1}
\end{table}
\begin{figure}[tbh]
    \centering
    \includegraphics[width=0.48\linewidth, height=0.35\linewidth]{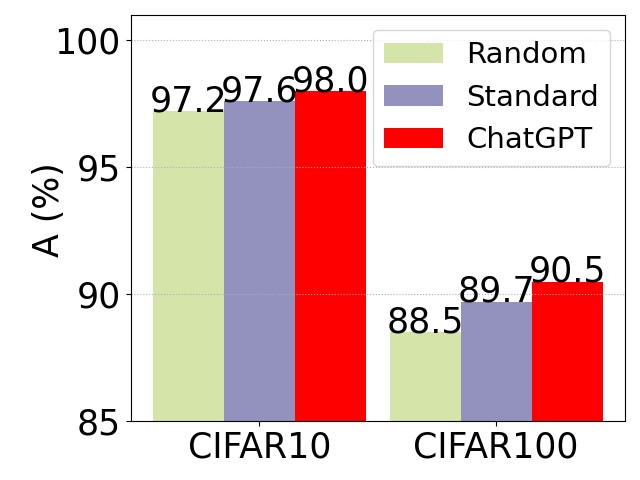}
    \includegraphics[width=0.48\linewidth, height=0.35\linewidth]{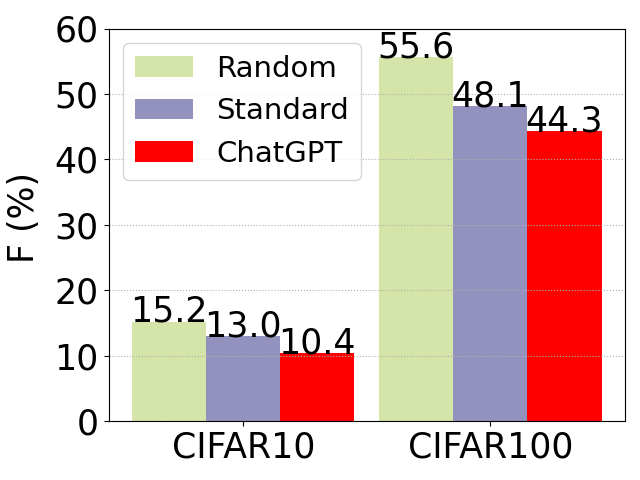}
    \caption{Ablation study of the text-guided learning on  CIFAR10 and CIFAR100 benchmarks with backbone ResNet-18. 
    All values are the average performance on the six OOD datasets. The proposed text-guided learning (`ChatGPT') is better than its two ablated versions.
    }
    \label{fig:ablation_text}
\end{figure}
\subsection{Ablation Study}
Extensive ablation studies were performed to confirm the effect of each component in the proposed learning framework. As
Table~\ref{tab:ablation1} shows on two benchmarks CIFAR100 and ImageNet100-I, when only one of the three main components (fake OOD data and two loss terms $\mathcal{L}_{CI}$ and $\mathcal{L}_{SC}$) is available, the model performs  better than the baseline without any of the three components (i.e., rows 2-4 vs. row 1). {Combination of any two components often leads to increased performance (rows 5-7) and already surpasses the best baseline on the CIFAR100 benchmark (AUROC 89.57\%-89.82\% vs. 87.43\%)}. Inclusion of all three components achieves the new state-of-the-art performance (last row), demonstrating the complementarity of the three components in improving OOD detection performance. 
\begin{figure*}[tbh]
    \centering
    \includegraphics[width=0.3\linewidth, height=0.23\linewidth]{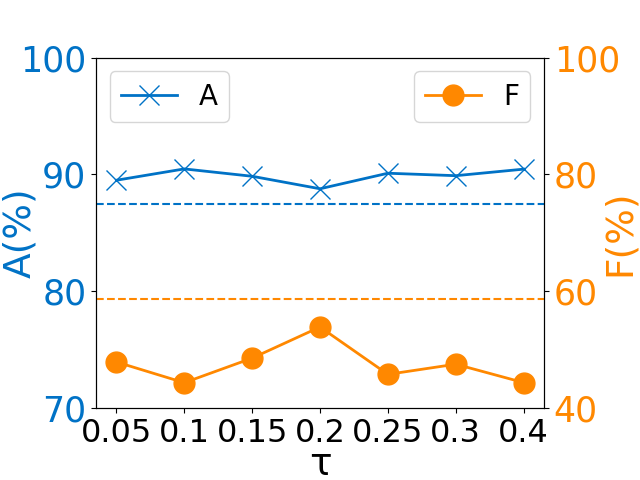}
    \includegraphics[width=0.3\linewidth, height=0.23\linewidth]{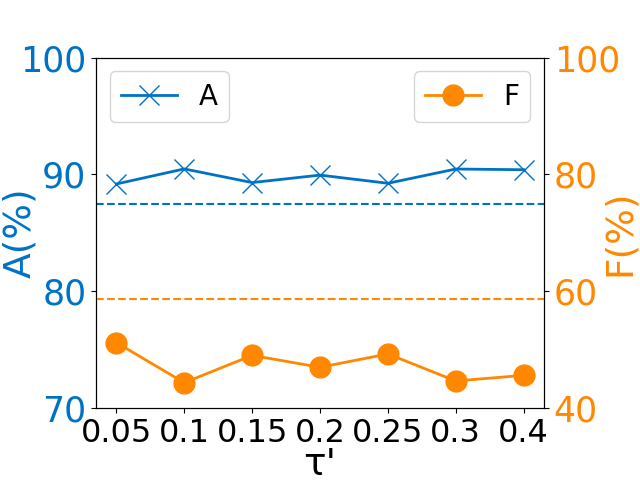}
    \includegraphics[width=0.3\linewidth, height=0.23\linewidth]{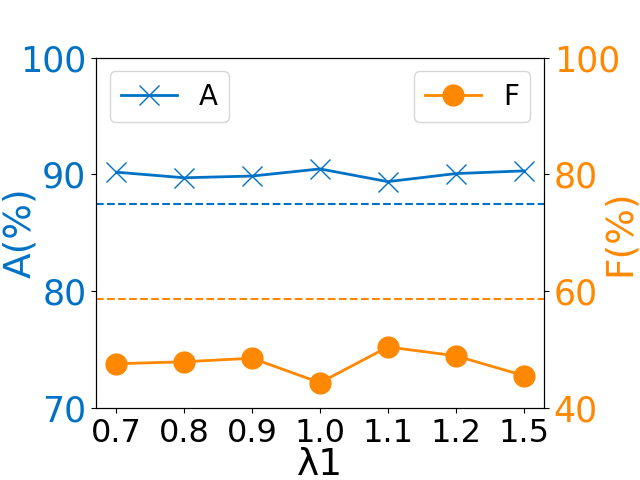}
    \includegraphics[width=0.3\linewidth, height=0.23\linewidth]{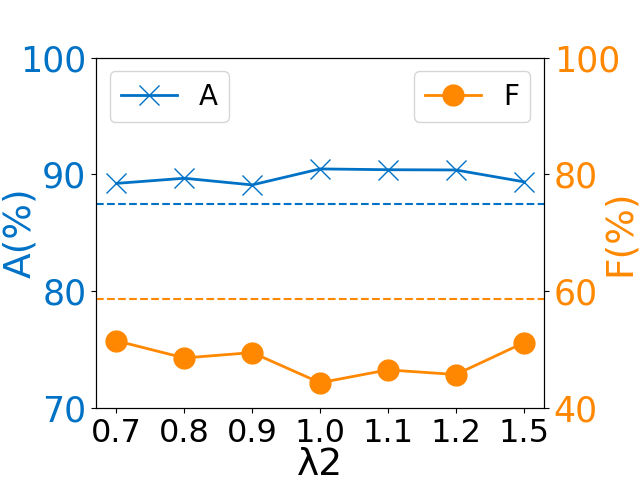}
    \includegraphics[width=0.3\linewidth, height=0.23\linewidth]{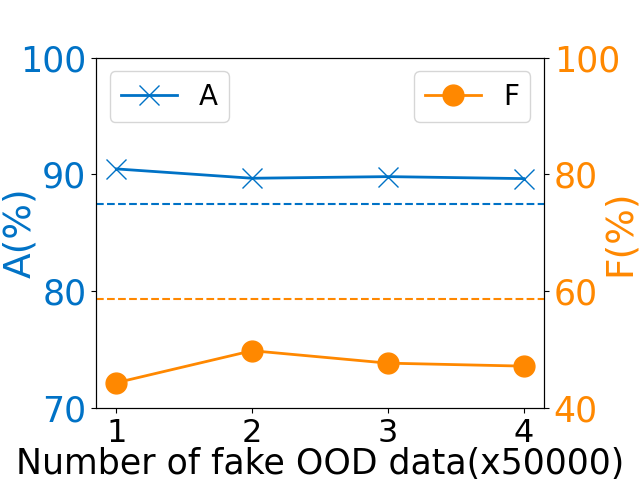}
    \includegraphics[width=0.3\linewidth, height=0.23\linewidth]{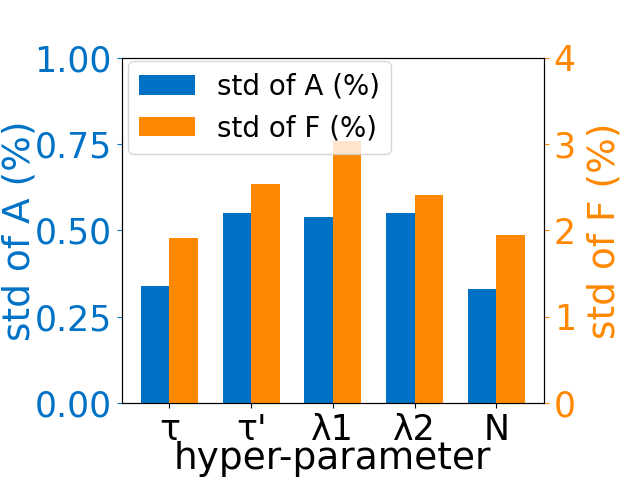}
    \caption{Sensitivity study of hyper-parameters $\tau$ and $\tau'$, $\lambda_1$ and $\lambda_2$,  and the number of fake OOD data. All experiments are on the CIFAR100 benchmark with model backbone ResNet18. The dashed line indicates the performance of the best baseline. Last subfigure: y-axis represents the standard deviation (std) of performance (A and F), x-axis represents five hyper-parameters, where N represents the number of fake OOD data.   
    }
    \label{fig:sensitivity}
\end{figure*}

In addition, more detailed ablation study on the text-guided learning was performed. Specifically, when the proposed ChatGPT-generated textual description (
Figure~\ref{fig:ablation_text}, 
{``ChatGPT''}) was replaced by the traditional simple description of each class (
{``Standard''}) in the form of ``a photo of [ID class name]'', or the ChatGPT-based anchor embedding was replaced by a randomly generated embedding (
{``Random''}) for each ID class, OOD detection performance was clearly downgraded on both CIFAR benchmarks (Figure~\ref{fig:ablation_text}). Similar results were obtained on the ImageNet100 benchmark (Supplementary Figure 2). This supports that both ChatGPT's textual description and CLIP's textual embeddings as anchors are helpful in guiding the learning of image encoder for OOD detection. Additional experiments on the validity of text selection and visualizations of more compact visual representations obtained from the proposed learning framework were in Supplementary Section E.
\subsection{Sensitive and Generalizability Studies}
The proposed learning framework is insensitive to the choice of hyper-parameters in a large range, including the temperature factors $\tau$ and $\tau'$, the weighting coefficients $\lambda_1$ and $\lambda_2$ in the loss function, and the number of fake OOD data used for model training. As Figure~\ref{fig:sensitivity} demonstrates, {when $\tau$ and $\tau'$ vary in the range $[0.05, 0.4]$, $\lambda_1$  and $\lambda_2$ in the range $[0.7, 1.5]$, $\lambda_2$ in the range $[0.7, 1.5]$, the number of fake OOD data in the range $[1\times 50,000, 4\times 50,000]$ (i.e., generating 1 to 4 fake OOD images for each of the 50,000 ID images), 
the model performs stably (as shown in Figure \ref{fig:sensitivity}: last subfigure representing the standard deviation (std) of performance on all hyper-parameters) and is better than the best baseline ReAct in AUROC, CIDER in FPR95 on the CIFAR100 benchmark. Similar results were obtained on the other benchmarks (Supplementary Figure 3).}

\begin{table}[ht]
\centering
\setlength{\tabcolsep}{3pt}
\small
\begin{tabular}{lccccc}
\hline
\multirow{2}{*}{Method} 
 & \multicolumn{2}{c}{ResNet50} & \multicolumn{2}{c}{ResNet101} \\ \cline{2-5}
 & F$\downarrow$ & A$\uparrow$ & F$\downarrow$ & A$\uparrow$\\ \hline
MSP & 58.54/\textbf{43.99}      & 87.92/\textbf{91.99}      & 55.56/\textbf{44.23}     & 88.70/\textbf{91.85}\\
Energy & 46.72/\textbf{38.22}      & 91.12/\textbf{92.78}      & 43.82/\textbf{36.57}     & 91.87/\textbf{93.35}      
                               \\
ViM                     & 57.97/\textbf{38.89}      & 88.94/\textbf{93.99}      & \textbf{51.21}/52.46     & 90.78/\textbf{91.53}   \\ 
ReAct                     & 39.85/\textbf{29.50}      & 92.12/\textbf{94.67}      & 39.98/\textbf{28.53}     & 92.26/\textbf{94.66}  \\\hline
\end{tabular}%
\caption{Fusion of our learning framework with various post-hoc OOD methods on the ImageNet100-I Benchmark. For each paired values by `/': the left one is from the original baseline and the right one is from the fusion one. Values are average percentages over four OOD datasets.}
\label{tab:imagenetfusion}
\end{table}
A further benefit of our learning framework is its flexible fusion with existing post-hoc OOD detection methods, where the proposed score function in the post-hoc methods are simply adopted for OOD detection after model training with our learning framework. 
Table~\ref{tab:imagenetfusion} and Table~\ref{tab:cifar_fusion} show that the fusion of our learning framework with each representative post-hoc method often improves the OOD detection performance compared to the original method on both the ImageNet100 and CIFAR benchmarks.

\begin{table*}[!bht]
\centering
\setlength{\tabcolsep}{4pt}
\normalsize
\begin{tabular}{lccccccccccc}
\hline
\multirow{3}{*}{Method} & \multicolumn{4}{c}{ResNet18} &  & \multicolumn{4}{c}{ResNet34} & \\ \cline{2-5} \cline{7-10} 
 & \multicolumn{2}{c}{CIFAR10} & \multicolumn{2}{c}{CIFAR100} &  & \multicolumn{2}{c}{CIFAR10} & \multicolumn{2}{c}{CIFAR100}&\\ \cline{2-5} \cline{7-10} 
 & F$\downarrow$ & A$\uparrow$ & F$\downarrow$ & A$\uparrow$ &  & F$\downarrow$ & A$\uparrow$ & F$\downarrow$ & A$\uparrow$&  \\ \hline
MSP & 48.03/\textbf{26.57}      & 91.40/\textbf{96.19}      & 73.60/\textbf{58.24}     & 82.24/\textbf{85.63}       &   & 40.95/\textbf{26.52}      & 92.09/\textbf{96.05}      & 78.29/\textbf{59.26}      & 79.25/\textbf{85.82} &           \\
Energy & 30.96/\textbf{10.51}      & 92.05/\textbf{97.95}      & 70.53/\textbf{47.42}     & 84.34/\textbf{89.46} &    & 26.69/\textbf{11.29}       
                               & 93.17/\textbf{97.64}      & 69.41/\textbf{53.30}      & 83.64/\textbf{88.82}&      \\
ViM                      & 47.90/\textbf{20.61}      & 91.02/\textbf{96.57}      & \textbf{71.43}/72.50     & \textbf{82.09}/81.76  &   & 38.35/\textbf{11.85}      
                               & 93.75/\textbf{97.79}      & 61.51/\textbf{55.19}      & 85.00/\textbf{87.95} &      \\ 
ReAct                     & 31.65/\textbf{10.41}      & 92.26/\textbf{97.99}      & 63.06/\textbf{44.33}     & 87.43/\textbf{90.47}   &  & 27.76/\textbf{11.17}       
                               & 93.29/\textbf{97.72}      & 50.56/\textbf{45.28}       & 88.30/\textbf{90.20} &     \\\hline
\end{tabular}%
\caption{Fusion of our framework with various OOD methods on the CIFAR Benchmarks. Values are average percentages over six OOD datasets.
}
\label{tab:cifar_fusion} 
\end{table*}

\section{Related Work}
OOD detection methods can be categorized into the following three groups based on accessible extra data and models.

\noindent\textbf{In-distribution data only:}
OOD detection methods with only ID data can be divided into two categories. One is training-based approach that incorporates regularization during model training~\cite{Mos,CSI,unsupood}, and the other is post-hoc approach which performs post-processing or additional analysis on the generally trained model to capture the discrepancy between ID and OOD data without model retraining.
For example, the training-based method G-ODIN~\cite{G-ODIN} {uses a divisor/dividend structure to measure the anomaly degree of input data}, and LogitNorm~\cite{LogitNorm} normalizes logits before the cross-entropy loss to reduce overconfidence. CIDER~\cite{CIDER} uses prototype construction during training, via which data from the same ID class become more compact and close to the associated ID-specific prototype, 
achieving best performance on the CIFAR10 benchmarks. In contrast, our approach uses semantically rich CLIP text embeddings as prototypes and achieves better performance on multiple benchmarks.

Differently in post-hoc methods, OOD scores are designed often based on information from top layers of generally trained neural networks, like softmax outputs~\cite{MSP, softmax1, softmax2}, logits~\cite{MaxLogits,Multi-label-Energy,Energy}, gradients~\cite{gradient,ODIN}, feature embeddings~\cite{Maha,Vim,featureembedding1,featureembedding2, FeatureNorm}, and model weights~\cite{DICE,LINe}. Our approach uses the state-of-the-art ReAct score~\cite{React} which improves the effect of the energy score by pruning high-activation feature components from the penultimate layer.

\noindent\textbf{Extra real or fake OOD data:} OOD detection performance is often improved when additional OOD data is accessible~\cite{oe_1,oe_2}. However, acquiring real OOD data is usually expensive. As an alternative solution, generating fake OOD data for OOD detection becomes popular and economically friendly. GANs have been used to generate synthetic OOD data~\cite{gan1,gan2,gan6}, but struggling with generation of complicated images and unstable training. VOS~\cite{VOS} assumes Gaussian-like feature distributions to synthesize outliers, while FeatureNorm~\cite{FeatureNorm} uses input-level fakes to find the layer in the pre-trained network with the largest difference in feature norm between ID and ODD data for OOD score design. 
Differently, our approach uses a simple yet effective Jigsaw strategy to generate challenging OOD data which are locally similar to but globally different from real ID data for model training, without complex generation process or extra assumptions. 

\noindent\textbf{Extra-modality model:} 
Recently, large vision-language models such as CLIP~\cite{CLIP} and ALIGN~\cite{align} have enabled major advancements in cross-modality studies. However, their usage as auxiliary tools for OOD detection remains limited. Fort et al.~\cite{CLIP_morelabel} send extra OOD text not included in ID classes to CLIP's text encoder for OOD detection. 
ZOC~\cite{CLIP_lablegenerate} train a label generator on CLIP's visual encoder to guide OOD detection, and similarly Ming et al.~\cite{MCM} design an OOD score based on the CLIP's visual and text encoders. However, all these studies require additional OOD labels and visual encoders.  Unlike these studies, our approach does not need any extra OOD label and CLIP's visual encoder.

\section{Conclusion}
In this study, a novel learning framework was proposed for OOD detection by using the Jigsaw-based fake OOD data and text-guided learning. The specially designed fake OOD data generation and the ChatGPT-based CLIP embedding for each ID class help the image encoder learn to extract more compact and semantic feature representation, which in turn helps discriminate between ID and OOD data as supported in extensive empirical evaluations. The new state-of-the-art performance of the proposed learning framework was obtained on the widely used benchmarks. Its flexible fusion with post-hoc methods indicates that the  learning framework may be easily combined with various new methods in future.

\section{Acknowledgements}
This work is supported in part by the Major Key Project of PCL (grant No. PCL2023AS7-1), the National Natural Science Foundation of China (grant No. 62071502), and Guangdong Excellent Youth Team Program (grant No. 2023B1515040025).

\bibliography{aaai24}

\clearpage

\end{document}